\documentclass{article}
\usepackage{spconf,amsmath,graphicx}
\usepackage{bm}
\usepackage{tabularx} % in the preamble
\usepackage{multirow,url}

\title{Motion Feature Augmented Recurrent Neural Network for Skeleton-based Dynamic Hand Gesture Recognition}
\name{Xinghao Chen \qquad Hengkai Guo \qquad Guijin Wang$^{\star}$ \qquad Li Zhang\thanks{This work was partially supported by National Science Foundation of China (No. 61271390) and State High-Tech R\&D Program of China (No. 2015AA016304). We gratefully acknowledge NVIDIA for GPU donation.}\thanks{$^{\star}$Corresponding author: wangguijin@tsinghua.edu.cn}}
\address{Department of Electronic Engineering, Tsinghua University, Beijing, China}
\begin{document}
%\ninept
%
\maketitle
\begin{abstract}
Dynamic hand gesture recognition has attracted increasing interests because of its importance for human computer interaction. In this paper, we propose a new motion feature augmented recurrent neural network for skeleton-based dynamic hand gesture recognition.
%For a dynamic hand gesture sequence, the finger and global movements are two important clues that can help to distinguish the gestures.
Finger motion features are extracted to describe finger movements and global motion features are utilized to represent the global movement of hand skeleton. These motion features are then fed into a bidirectional recurrent neural network (RNN) along with the skeleton sequence, which can augment the motion features for RNN and improve the classification performance.
%The proposed method is evaluated on the challenging skeleton-based DHG-14/28 dataset.
Experiments demonstrate that our proposed method is effective and outperforms start-of-the-art methods.

\end{abstract}
\section{Introduction}
\label{sec:intro}
%% introduction to hand gesture recognition
Due to its flexibility and expressiveness, hand gesture can provide an efficient and natural way for human computer interaction (HCI). Hand gesture recognition has been researched for decades and has great potentials for applications in sign language recognition, remote control and virtual reality etc~\cite{mitra2007gesture,zeng2012hand,chen2016static,dong2015american,ohn2014hand,molchanov2016online,Neverova2016moddrop}.
%% introduction to skeleton-based method
%Generally, hand gesture recognition can be categorized into static hand gesture recognition~\cite{chen2016static,dong2015american} and dynamic hand gesture recognition~\cite{ohn2014hand,molchanov2016online,Neverova2016moddrop}. In this paper, we specifically focus on dynamic hand gesture recognition.
Dynamic hand gesture recognition aims to understand what a hand sequence conveys. It remains a challenging task due to high intra-class variance because the way of performing a gesture differs from person to person.

Previous works on dynamic hand gesture recognition usually took RGB images and depth images~\cite{wang2013depth,shi2015high} as input~\cite{ohn2014hand}. Some of them used multi-modal input including IR images~\cite{molchanov2016online} or audio stream~\cite{Neverova2016moddrop}.
%Since the great success of human action recognition from 3D skeleton sequence, the area of skeleton-based dynamic hand gesture recognition has achieved more attention.
Recent progresses on hand pose estimation~\cite{supancic2015depth,oberweger2015training,tang2015opening,YeSpatialHandECCV2016}
have greatly promoted the research on dynamic hand gesture recognition from 3D hand skeleton sequences.
Smedt et al.~\cite{de2016skeleton} proposed a skeleton-based approach for dynamic hand gesture recognition and demonstrated its superiority over depth-based approaches. In their approach, a temporal pyramid representation was utilized to model temporal information. Shape of connected joints, histogram of hand directions and wrist rotations were used to characterize hand shape and hand movement. However, the amplitude of gesture is not considered in their approach and the temporal pyramid representation may lose some motion information.
%However, a better way to model temporal information and more powerful features will help to improve the performances.
%~\cite{supancic2015depth,tompson2014real,oberweger2015training,tang2015opening,sun2015cascaded,YeSpatialHandECCV2016}
%Smedt et al.~\cite{de2016skeleton} proposed an algorithm to classify dynamic hand gestures from hand skeleton sequences. The shape of connected joints (SoCJ) features were extracted to describe the hand shape and a temporal pyramid method was utilized to model the temporal information of hand gestures.

\begin{figure}[t]
  \centering
  \centerline{\includegraphics[width=\linewidth]{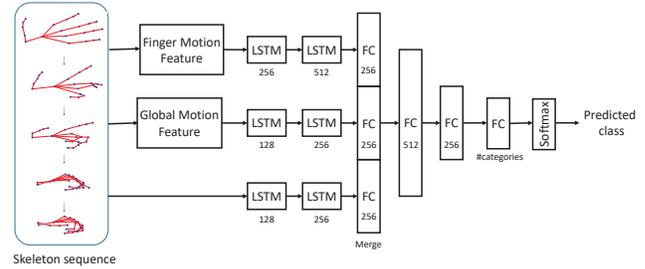}}
%  \vspace{2.0cm}
\caption{The framework of our proposed method. Finger motion features and global motion features are extracted from the input dynamic hand gesture skeleton sequence. These motion features, along with the skeleton sequence, are fed into a recurrent neural network (RNN) to get the predicted class of input gesture.}
% The RNN consists of two long-short term memory (LSTM) layers and three fully connected (FC) layers.
\label{fig:framework}
\end{figure}

%% introduction to features
%EigenJoints features~\cite{yang2012eigenjoints} are widely used in skeleton-based action Recognition.
The most important clues for dynamic hand gesture are articulated movements of fingers and the global movements of the hand.
In prior works, some sort of joint angle features~\cite{dong2015american, lu2016dynamic} were utilized to describe the hand shape.
%Lu et al.~\cite{lu2016dynamic} proposed single finger features and double finger features to distinguish dynamic hand gestures from Leapmotion controller.
However, these features are not sufficient enough to characterize the full pose of a hand.

%Inspired by these works,
In this paper, we propose a motion feature augmented RNN for skeleton-based dynamic hand gesture recognition. We extract the angles of bones from the hand skeleton, which is efficient and concise representation of the finger articulated movements. To describe the global movements of the hand, we extract the global rotation and global translation of the hand. A distance adaptive discretization scheme is given to better model the amplitude of the gestures. The finger motion features and global features are fed into a bidirectional RNN along with the skeleton sequence to predict the class of input gesture. Experiments on the publicly-available skeleton-based DHG-14/28 dataset~\cite{de2016skeleton} demonstrate the effectiveness of our proposed method.

%\section{Related work}
%\label{sec:relatedwork}
%
%dynamic and static hand gesture
%
%{\bf{Feature augmented method}}. Combining hand-crafted features and CNN features~\cite{wang2016combining}
%Feature Augmented Deep Neural Networks~\cite{sadanandan2016feature}
%
%{\bf{skeleton-based action recognition}}.
%EigenJoints-based Action Recognition~\cite{yang2012eigenjoints}.
%
%{\bf{Multi-modal input}}.
%
%single finger feature and double finger feature~\cite{lu2016dynamic}
%
%root-center angle~\cite{chen2016static}

\section{Proposed Framework}
\label{sec:framework}
The framework of our proposed algorithm is shown in Figure~\ref{fig:framework}. A hand skeleton sequence is taken as input and the class of gesture is predicted by RNN. Firstly the global motion features and finger motion features are extracted from the input skeleton sequence.
The hand skeleton can be directly and effectively represented by a kinematic hand model whose parameters are the angles of bones, the global translation and global rotation\cite{tang2015opening,YeSpatialHandECCV2016}. Therefore, these hand parameters can serve as efficient and discriminating features for dynamic hand gesture recognition. In our approach, theses features with offset and dynamic pose modelling are utilized as the motion features to represent dynamic hand gestures.
The details of motion feature extraction will be presented in Section~\ref{sec:feature}.
%% introduction to temporal modelling
%Recurrent neural network (RNN) is a widely used framework for temporal information modelling and has obtained great success in temporal sequences recognition task like speech recognition~\cite{graves2013speech}, action recognition~\cite{du2015hierarchical} and hand gesture recognition~\cite{molchanov2016online} etc.

We exploit the recurrent neural network (RNN) to model temporal information for its success in temporal sequences recognition tasks\cite{du2015hierarchical, molchanov2016online}.
%% introduction to feature augmented dnn
Though RNN can somehow learn features from the input sequences, some information may be absent or weakened, which will hinder the classification performance. Some previous works~\cite{wang2016combining,sadanandan2016feature} combined features extracted from deep neural network with hand-crafted features to enhance the discriminability of the features. Inspired by these works, in this paper we utilized a RNN which is augmented by motion features to classify dynamic hand gestures from skeleton sequence.
The finger motion features and global features are fed into a RNN along with the skeleton sequence to predict the class of input gesture, which will be discussed in Section~\ref{sec:dhgr}.

\section{Motion Feature Extraction}
\label{sec:feature}
\begin{table*}[!htb]
\centering
\caption{Recognition rates (\%) of self-comparison experiments on DHG-14 dataset.}
\label{tab:com_self}
\vspace{0.1cm}
\begin{tabular}{|*{10}{c|}}
\hline
\multirow{2}*{Method}
& \multicolumn{3}{c|}{fine} & \multicolumn{3}{c|}{coarse}& \multicolumn{3}{c|}{both}\\\cline{2-10}
 & best & worst & avg$\pm$std & best & worst & avg$\pm$std & best & worst & avg$\pm$std\\\hline
Skeleton & 86.0 & 42.0 & ${61.2\pm12.37}$ & $\bm{97.78}$ & $\bm{74.44}$ & ${86.44\pm7.94}$ & 93.57 & $\bm{67.86}$ & ${77.43\pm6.82}$\\\hline
Motion Features & 84.0 & 46.0 & $71.5\pm11.44$ & 96.67 & 64.44 & $81.94\pm8.17$ & 90.0 & 58.57 & $78.21\pm7.49$\\\hline
Ours & $\bm{90.0}$ & $\bm{56.0}$ & $\bm{76.9\pm9.19}$ & $\bm{97.78}$ & 72.22 & $\bm{89.0\pm7.55}$ & $\bm{94.29}$ & $\bm{67.86}$ & $\bm{84.68\pm6.67}$\\\hline
\end{tabular}
\end{table*}
%\begin{table*}[!htb]
%\centering
%\caption{Results of our proposed method for 28 gestures on the DHG dataset.}
%\label{tab:28gestureaccu}
%\begin{tabular}{|*{10}{c|}}
%\hline
%\multirow{2}*{Method}
%& \multicolumn{3}{c|}{fine} & \multicolumn{3}{c|}{coarse}& \multicolumn{3}{c|}{both}\\\cline{2-10}
% & best & worst & avg$\pm$std & best & worst & avg$\pm$std & best & worst & avg$\pm$std\\\hline
%DHG~\cite{de2016skeleton} & - & - & - & - & - & - & - & - & 80.0\\\hline
%Ours & 86.0 & 46.0 & 70.2$\pm$11.57 & 96.67 & 68.89 & 85.61$\pm$8.29 & 89.29 & 60.71 & $\bm{80.11\pm7.07}$ \\\hline
%\end{tabular}
%\end{table*}

%\begin{table*}[!htb]
%\centering
%\caption{Results of our proposed method for 28 gestures on the DHG dataset.}
%\label{tab:28gestureaccu}
%\begin{tabular}{|*{10}{c|}}
%\hline
%\multirow{2}*{Method}
%& \multicolumn{3}{c|}{fine} & \multicolumn{3}{c|}{coarse}& \multicolumn{3}{c|}{both}\\\cline{2-10}
% & best & worst & avg$\pm$std & best & worst & avg$\pm$std & best & worst & avg$\pm$std\\\hline
%Smedt et al.~\cite{de2016skeleton} & - & - & - & - & - & - & - & - & $80.0\pm-$ \\\hline
%Ours & 84.0 & 38.0 & 68.0$\pm$12.62 & 98.89 & 72.22 & 87.17$\pm$7.30 & 89.29 & 62.14 & $\bm{80.32\pm6.73}$ \\\hline
%\end{tabular}
%\end{table*}
In this section we will describe how to extract finger motion features $\mathcal{H}(\mathcal{S})$ and global motion features $\mathcal{G}(\mathcal{S})$ from the input hand skeleton sequence $\mathcal{S}=\{s^t\}_{t=1}^{T}$, where $s^t=\{x_i^t,y_i^t,z_i^t\}_{i=1}^J$ denotes a hand skeleton for frame $t$, $T$ is the number of frames of this sequence and $J$ is the number of joints for hand skeleton.

\subsection{Global Motion Feature}
\label{sec:globalfeature}
The global motion features (global rotation and global translation) are important for dynamic hand gesture. Typically, the global status of the hand can be represented by the wrist joint, palm joint and metacarpophalangeal (MCP) joints, which is denoted by $p^t$. We use Kabsch algorithm~\cite{kabsch1976solution} to infer the global rotation $\mathcal{G}_r$ and global translation $\mathcal{G}_l$, as shown in Equation~(\ref{eq:grl}):
\begin{equation}
\label{eq:grl}
[\mathcal{G}_l, \mathcal{G}_r] = \mathnormal{Kabsch}(p^t, p_0)
\end{equation}
where $\mathcal{G}_r = (r_x, r_y, r_z)$ represents the rotations along three axis and $\mathcal{G}_l = (\rho, \theta, \phi)$ is the spherical coordinates of global translation. $p_0$ is a fake palm that centers at $(0,0,0)$ and faces the camera.

The amplitudes of hand gestures differ from person to person for the same gesture. Therefore previous work~\cite{de2016skeleton} ignored the amplitude part $\rho$ of global translation. However, sometimes the amplitude is critical for gestures. For example, gesture $\it{Grab}$ and gesture $\it{Pinch}$ are quite similar except for the amplitude of the gesture. To this end, we propose a distance adaptive discretization(DAD) method to extract global translation amplitude feature, inspired by Distance Adaptive Scheme~\cite{liang2014parsing, dong2015american} which is used for feature selection. The DAD method discretizes $\rho$ into $M$ bins using the threshold $\{\eta_{i}\}_{i=1}^M$. A gaussian distribution kernel $g(x)$ is used to generate the thresholds.
\begin{equation}
\label{eq:gth}
\int_0^{\eta_i} g(x) dx = \frac {i} {M}\int_0^{\sigma} g(x) dx
\end{equation}
where $\sigma$ is the standard deviation of the gaussian function. In our experiments, $\sigma=1.5r_{palm}$ where $r_{palm}$ is the radius of the palm. The global feature can be written as Equation~(\ref{eq:gf}):
\begin{equation}
\label{eq:gf}
\Phi^t = [\rho_{bin}, \theta, \phi, r_x, r_y, r_z]
\end{equation}
where $\rho_{bin}$ is the discrete representation of $\rho$ using the thresholds determined by Equation~(\ref{eq:gth}).

Similarly to previous works~\cite{chen2016novel}, we use offset pose $\Phi_{op}^t$ and dynamic pose $\Phi_{dp}^t$ to model the finger motion features. The offset pose represents the offset from current pose to the pose of first frame of the gesture sequence. The dynamic pose represents the difference of global features between current frame and several previous frames. There features can enhance the representability of the global motion of the hand and thus can model the temporal information of dynamic hand gesture.
\begin{equation}
\label{eq:gop}
\Phi_{op}^t = \Phi^t - \Phi^1
\end{equation}
\begin{equation}
\label{eq:gdp}
\Phi_{dp}^t = \{\Phi^t - \Phi^{t-s}|s=1,5,10\}
\end{equation}
All above features are concatenated to form the global motion features $\mathcal{G}^t(\mathcal{S})=[\Phi^t, \Phi_{op}^t, \Phi_{dp}^t]$ for frame $t$.

\subsection{Finger Motion Feature}
\label{sec:fingerfeature}
For many dynamic hand gesture, the movement of fingers are critical because the global movement may be non-significant, especially for fine-grained gestures.
%Previous work either use the ~\cite{de2016skeleton}
%We use 26 DoFs (degree of freedoms) to model the hand. The global translation and global orientation has 6 DoFs, which represent the global status of the hand.
We use 20 DoFs (degree of freedoms) to model the finger movement. For the MCP joints, there are 2 DoFs for each joint. For proximal interphalangeal (PIP) and distal interphalangeal (DIP) joints, 1 DoF is used to describe the angle of bone. These parameters can retain rich information for the shape of the hand skeleton. We use $\mathcal{IK}(\cdot)$ to denote the inverse kinematics function that derive hand parameters from the original hand skeleton $s^t$.
\begin{equation}
\label{eq:ik}
\Theta^t = \mathcal{IK}(s^t)
\end{equation}
Similarly, we use dynamic pose $\Theta_{dp}^t$ and offset pose $\Theta_{op}^t$ to model the finger motion feature.
\begin{equation}
\label{eq:fop}
\Theta_{op}^t = \Theta^t - \Theta^1
\end{equation}
\begin{equation}
\label{eq:fdp}
\Theta_{dp}^t = \{\Theta^t - \Theta^{t-s} | s=1,5,10\}
\end{equation}
These features are concatenated to form the finger motion features $\mathcal{F}^t(\mathcal{S})=[\Theta^t, \Theta_{op}^t, \Theta_{dp}^t]$ for frame $t$.

\section{Dynamic Hand Gesture Recognition}
\label{sec:dhgr}

RNN has shown great successes in human action recognition and hand gesture recognition. Although RNN can learn features for the input data, the representability of the features may be absent in some aspects. To this end, we augment features for RNN by combining the hand-crafted global and finger motion features and the original skeleton. The framework of our proposed method for skeleton-based dynamic hand gesture recognition is shown in Figure~\ref{fig:framework}. The finger motion features and global motion features are extracted from the input skeleton sequence. These motion features and the input skeleton sequence are fed into the RNN. Each branch contains two long short term memory (LSTM) layers and one fully connected (FC) layer. Outputs from three branches are concatenated together, followed by three FC layers and a softmax layer for class prediction. All layers are followed by a dropout layer and FC layers are followed by a ReLU function.

\section{Experiments}
\label{sec:exp}
%In this section we will show the experimental results of our proposed method. Firstly the dataset used for the experiments and some implementation details are briefly introduction. Secondly the performance on 14 gestures and 28 gestures are shown and discussed.
\begin{figure*}[htb]
  \centering
  \centerline{\includegraphics[width=0.93\linewidth]{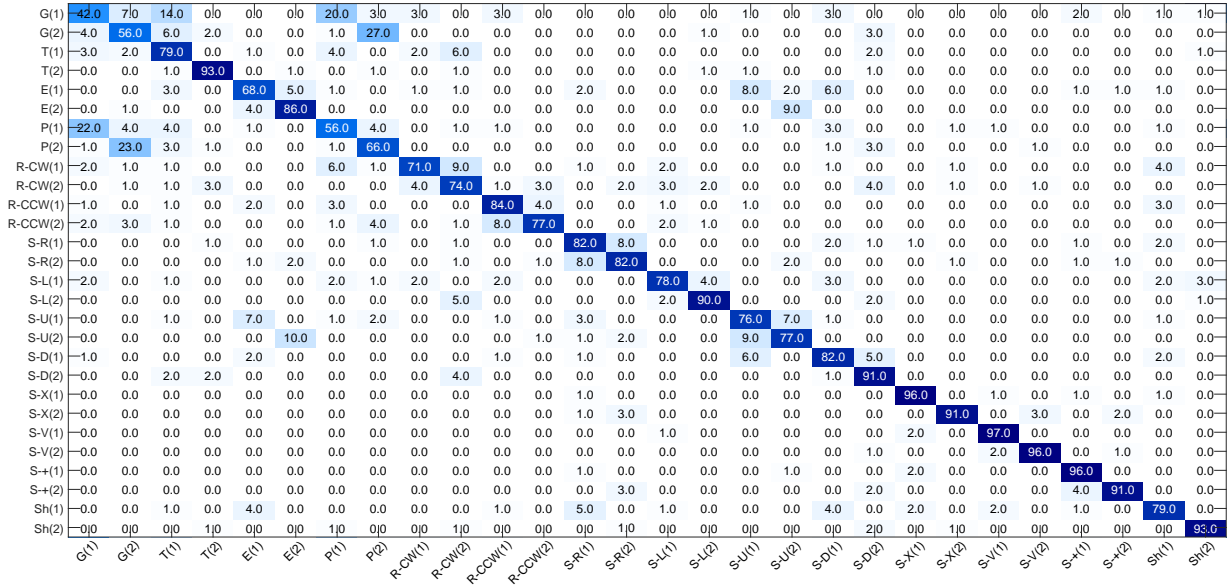}}
%  \vspace{2.0cm}
\caption{The confusion matrix of the proposed approach for DHG-28.}
\label{fig:res28}
\end{figure*}
\subsection{Dataset}
\label{ssec:dataset}

DHG-14/28~\cite{de2016skeleton} is a public dynamic hand gesture dataset that provides hand gesture sequences with depth images and skeletons. Since our proposed method bases on hand skeleton, we only use the skeleton information of the dataset to conduct our experiments. DHG-14/28 is a challenging dataset since it contains hand gesture from 20 subjects and has 14 gestures with two different finger configurations.

\subsection{Implementation}
The proposed RNN framework is implemented in Keras~\cite{chollet2015keras}. We use Adam~\cite{kingma2014adam} algorithm with mini-batch of 32 to train the network. The parameters of Adam are set to default setting suggested in~\cite{kingma2014adam}, with learning $lr=0.001$, $\beta_1=0.9$, $\beta_2=0.999$ and $\epsilon=1e^{-08}$.The network is trained for 100 epochs. In our experiments, $M$ of Equation~(\ref{eq:gth}) is set to $M=5$. Every skeleton sequence is subtracted by the palm position of the first frame and scaled the amplitude to $1$ before fed into third branch in Figure~\ref{fig:framework}.

\subsection{Self-comparison}
\label{ssec:14gesture}
To verify the contributions of our proposed method, we conduct two self-comparison baseline experiments on DHG-14 dataset, which has 14 gesture classes. The first baseline (Motion Features) only takes motion features as input and remove the third branch of the framework shown in Figure~\ref{fig:framework}. The second baseline (Skeleton) only use the skeleton sequences as input.
\begin{figure}[htb]
  \centering
  \centerline{\includegraphics[width=\linewidth]{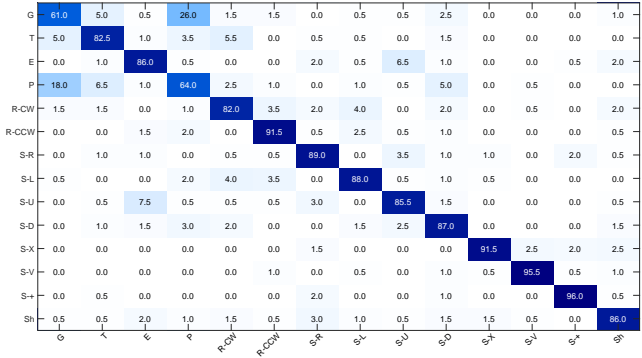}}
%  \vspace{2.0cm}
\caption{The confusion matrix of the proposed approach for DHG-14.}
\label{fig:res14}
\end{figure}
We follow same experimental setup as~\cite{de2016skeleton}, using a leave-one subject-out cross-validation (LOOCV) strategy for all following experiments. The proposed network is trained on data from 19 subjects and tested on the rest one. Therefore, these experiments are repeated 20 times, with different subject being used for testing.
Previous work~\cite{de2016skeleton} only reported the average classification accuracy, which is not sufficient to evaluate the performance of the algorithm. In this paper, we report the worst, best and average results of 20 different splitting protocol as well as the standard derivation.

The recognition rates of these two baselines are shown in Table~\ref{tab:com_self}. In most cases, Our proposed method outperforms two baselines in terms of worst, best, average accuracy and the stand derivation, which verify the effectiveness of the proposed framework.
\begin{table}[!htb]
\centering
\caption{Comparison of recognition rates (\%) on DHG-14/28 dataset.}
\label{tab:com_prior}
\vspace{0.1cm}
\begin{tabular}{|*{5}{c|}}
\hline
\multirow{2}*{Method}
& \multicolumn{3}{c|}{DHG-14} & \multicolumn{1}{c|}{DHG-28}\\\cline{2-5}
 & fine & coarse & both & both\\\hline
Smedt et al.~\cite{de2016skeleton} & 73.60 & 88.33 & 83.07 & 80.0 \\\hline
Ours & $\bm{76.9}$ & $\bm{89.0}$ & $\bm{84.68}$ & $\bm{80.32}$ \\\hline
\end{tabular}
\end{table}
\subsection{Comparison with State-of-the-arts}
\label{sssec:28gesture}

We compare our work with state-of-the art method~\cite{de2016skeleton} on DHG-14/28 dataset. The recognition rates of different methods on DHG-14 and DHG-28 dataset are shown in Table~\ref{tab:com_prior}. It shows that our proposed method outperforms state-of-the-art work~\cite{de2016skeleton} on DHG-14 dataset in terms of coarse, fine and all gestures. %worst, best, average accuracy and the stand derivation.
To better illustrate the performance of our proposed algorithm, the confusion matrix of 14 classes is shown in Figure~\ref{fig:res14}. It can be observed that the confusion between gesture {\it{Grab}} and {\it{Pinch}} is severe, due to the high similarity of these two gestures. However, our algorithm does improve the performance of these two gestures compared with those of ~\cite{de2016skeleton}. It can be observed that our method promotes the classification accuracy of fine-grained gestures a lot. The improvement of recognition rate of coarse-grained gestures is comparatively little because it's already quite good.

As shown in Table~\ref{tab:com_prior}, our method is also better than ~\cite{de2016skeleton} when considering the more complicated 28-gestures classification task, which demonstrates the effectiveness of our proposed algorithm.
The confusion matrix of 28 classes is shown in Figure~\ref{fig:res28}.
\begin{table}[!htb]
\centering
\caption{Comparison of LAFED metric.}
\vspace{0.1cm}
\label{tab:lafed}
\begin{tabular}{|c|c|c|}
\hline
Method & Smedt et al.~\cite{de2016skeleton} & Ours \\\hline
LAFED & 0.0114 & $\bm{0.0075}$ \\\hline
\end{tabular}
\end{table}
%\subsection{LAFED metric}
%\label{sec:lafed}
A metric called Loss of Accuracy when Removing the Finger Differentiation (LARFD) was proposed in~\cite{de2016skeleton} to evaluate to what degree we can blame the loss of accuracy from 14-gestures to 28-gestures classification on the intra-gesture confusion. The smaller LARFD metric is, the less loss of accuracy is due to intra-gesture confusion. The LARFD metric of different methods is listed in Table~\ref{tab:lafed}. We can see that our proposed algorithm outperforms ~\cite{de2016skeleton}.

\section{Conclusion}
\label{sec:conclusion}

This paper proposes an algorithm to augment the motion features for recurrent neural network to recognize skeleton-based dynamic hand gestures. The finger motion features are extracted from the input skeleton sequence to describe the articulated movements of fingers and the global motion features are extracted to represent the global translation and rotation of the hand. The motion features, along with the skeleton sequence, are fed into a RNN to predict the class of input gesture. Experiments on the public DHG-14/28 dataset demonstrate that our proposed method outperforms state-of-the-art methods. Future work may focus on a hierarchical coarse to fine framework to achieve better classification performance.

%\section{Acknowledgements}
%\label{sec:ack}

% Below is an example of how to insert images. Delete the ``\vspace'' line,
% uncomment the preceding line ``\centerline...'' and replace ``imageX.ps''
% with a suitable PostScript file name.
% -------------------------------------------------------------------------

% To start a new column (but not a new page) and help balance the last-page
% column length use \vfill\pagebreak.
% -------------------------------------------------------------------------
%\vfill
%\pagebreak

% References should be produced using the bibtex program from suitable
% BiBTeX files (here: strings, refs, manuals). The IEEEbib.bst bibliography
% style file from IEEE produces unsorted bibliography list.
% -------------------------------------------------------------------------
\bibliographystyle{IEEEbib}
\bibliography{refs}

\begin{thebibliography}{10}

\bibitem{mitra2007gesture}
Sushmita Mitra and Tinku Acharya,
\newblock ``Gesture recognition: A survey,''
\newblock {\em IEEE Transactions on Systems, Man, and Cybernetics, Part C
  (Applications and Reviews)}, vol. 37, no. 3, pp. 311--324, 2007.

\bibitem{zeng2012hand}
Bobo Zeng, Guijin Wang, and Xinggang Lin,
\newblock ``A hand gesture based interactive presentation system utilizing
  heterogeneous cameras,''
\newblock {\em Tsinghua Science and Technology}, vol. 17, no. 3, pp. 329--336,
  2012.

\bibitem{chen2016static}
Xinghao Chen, Chenbo Shi, and Bo~Liu,
\newblock ``Static hand gesture recognition based on finger root-center-angle
  and length weighted mahalanobis distance,''
\newblock in {\em SPIE Photonics Europe}. International Society for Optics and
  Photonics, 2016, pp. 98970U--98970U.

\bibitem{dong2015american}
Cao Dong, Ming~C Leu, and Zhaozheng Yin,
\newblock ``American sign language alphabet recognition using microsoft
  kinect,''
\newblock in {\em Proceedings of the IEEE Conference on Computer Vision and
  Pattern Recognition Workshops}, 2015, pp. 44--52.

\bibitem{ohn2014hand}
Eshed Ohn-Bar and Mohan~Manubhai Trivedi,
\newblock ``Hand gesture recognition in real time for automotive interfaces: A
  multimodal vision-based approach and evaluations,''
\newblock {\em IEEE Transactions on Intelligent Transportation Systems}, vol.
  15, no. 6, pp. 2368--2377, 2014.

\bibitem{molchanov2016online}
Pavlo Molchanov, Xiaodong Yang, Shalini Gupta, Kihwan Kim, Stephen Tyree, and
  Jan Kautz,
\newblock ``Online detection and classification of dynamic hand gestures with
  recurrent 3d convolutional neural network,''
\newblock in {\em Proceedings of the IEEE Conference on Computer Vision and
  Pattern Recognition}, 2016, pp. 4207--4215.

\bibitem{Neverova2016moddrop}
N.~Neverova, C.~Wolf, G.~Taylor, and F.~Nebout,
\newblock ``Moddrop: Adaptive multi-modal gesture recognition,''
\newblock {\em IEEE Transactions on Pattern Analysis and Machine Intelligence},
  vol. 38, no. 8, pp. 1692--1706, Aug 2016.

\bibitem{wang2013depth}
Guijin Wang, Xuanwu Yin, Xiaokang Pei, and Chenbo Shi,
\newblock ``Depth estimation for speckle projection system using progressive
  reliable points growing matching,''
\newblock {\em Applied optics}, vol. 52, no. 3, pp. 516--524, 2013.

\bibitem{shi2015high}
Chenbo Shi, Guijin Wang, Xuanwu Yin, Xiaokang Pei, Bei He, and Xinggang Lin,
\newblock ``High-accuracy stereo matching based on adaptive ground control
  points,''
\newblock {\em IEEE Transactions on Image Processing}, vol. 24, no. 4, pp.
  1412--1423, 2015.

\bibitem{supancic2015depth}
James~S Supancic, Gr{\'e}gory Rogez, Yi~Yang, Jamie Shotton, and Deva Ramanan,
\newblock ``Depth-based hand pose estimation: data, methods, and challenges,''
\newblock in {\em Proceedings of the IEEE International Conference on Computer
  Vision}, 2015, pp. 1868--1876.

\bibitem{oberweger2015training}
Markus Oberweger, Paul Wohlhart, and Vincent Lepetit,
\newblock ``Training a feedback loop for hand pose estimation,''
\newblock in {\em Proceedings of the IEEE International Conference on Computer
  Vision}, 2015, pp. 3316--3324.

\bibitem{tang2015opening}
Danhang Tang, Jonathan Taylor, Pushmeet Kohli, Cem Keskin, Tae-Kyun Kim, and
  Jamie Shotton,
\newblock ``Opening the black box: Hierarchical sampling optimization for
  estimating human hand pose,''
\newblock in {\em Proceedings of the IEEE International Conference on Computer
  Vision}, 2015, pp. 3325--3333.

\bibitem{YeSpatialHandECCV2016}
Qi~Ye, Shanxin Yuan, and Tae-Kyun Kim,
\newblock ``Spatial attention deep net with partial pso for hierarchical hybrid
  hand pose estimation,''
\newblock in {\em The European Conference on Computer Vision (ECCV)}, 2016.

\bibitem{de2016skeleton}
Quentin De~Smedt, Hazem Wannous, and Jean-Philippe Vandeborre,
\newblock ``Skeleton-based dynamic hand gesture recognition,''
\newblock in {\em Proceedings of the IEEE Conference on Computer Vision and
  Pattern Recognition Workshops}, 2016, pp. 1--9.

\bibitem{lu2016dynamic}
Wei Lu, Zheng Tong, and Jinghui Chu,
\newblock ``Dynamic hand gesture recognition with leap motion controller,''
\newblock {\em IEEE Signal Processing Letters}, vol. 23, no. 9, pp. 1188--1192,
  2016.

\bibitem{du2015hierarchical}
Yong Du, Wei Wang, and Liang Wang,
\newblock ``Hierarchical recurrent neural network for skeleton based action
  recognition,''
\newblock in {\em Proceedings of the IEEE Conference on Computer Vision and
  Pattern Recognition}, 2015, pp. 1110--1118.

\bibitem{wang2016combining}
Pichao Wang, Zhaoyang Li, Yonghong Hou, and Wanqing Li,
\newblock ``Combining convnets with hand-crafted features for action
  recognition based on an hmm-svm classifier,''
\newblock {\em arXiv preprint arXiv:1602.00749}, 2016.

\bibitem{sadanandan2016feature}
Sajith~Kecheril Sadanandan, Petter Ranefall, and Carolina W{\"a}hlby,
\newblock ``Feature augmented deep neural networks for segmentation of cells,''
\newblock in {\em European Conference on Computer Vision Workshops}. Springer,
  2016, pp. 231--243.

\bibitem{kabsch1976solution}
Wolfgang Kabsch,
\newblock ``A solution for the best rotation to relate two sets of vectors,''
\newblock {\em Acta Crystallographica Section A: Crystal Physics, Diffraction,
  Theoretical and General Crystallography}, vol. 32, no. 5, pp. 922--923, 1976.

\bibitem{liang2014parsing}
Hui Liang, Junsong Yuan, and Daniel Thalmann,
\newblock ``Parsing the hand in depth images,''
\newblock {\em IEEE Transactions on Multimedia}, vol. 16, no. 5, pp.
  1241--1253, 2014.

\bibitem{chen2016novel}
Hongzhao Chen, Guijin Wang, Jing-Hao Xue, and Li~He,
\newblock ``A novel hierarchical framework for human action recognition,''
\newblock {\em Pattern Recognition}, vol. 55, pp. 148--159, 2016.

\bibitem{chollet2015keras}
Fran\c{c}ois Chollet,
\newblock ``Keras,'' \url{https://github.com/fchollet/keras}, 2015.

\bibitem{kingma2014adam}
Diederik Kingma and Jimmy Ba,
\newblock ``Adam: A method for stochastic optimization,''
\newblock {\em arXiv preprint arXiv:1412.6980}, 2014.

\end{thebibliography}

\end{document}